# A Novel End-To-End Event Geolocation Method Leveraging Hyperbolic Space and Toponym Hierarchies

Yaqiong Qiao[a,*], Guojun Huang[b]

[a] Nankai University, Tianjin, China 450046

[b] School of Information Engineering, North China University of Water Resources and Electric Power, China 450046

Abstract:

Timely detection and geolocation of events based on social data can provide critical information for applications such as crisis response and resource allocation. However, most existing methods are greatly affected by event detection errors, leading to insufficient geolocation accuracy. To this end, this paper proposes a novel end-to-end event **g**eolocation method (GTOP) leveraging Hyperbolic space and **top**onym hierarchies. Specifically, the proposed method contains one event detection module and one geolocation module. The event detection module constructs a heterogeneous information networks based on social data, and then constructs a homogeneous message graph and combines it with the text and time feature of the message to learning initial features of nodes. Node features are updated in Hyperbolic space and then fed into a classifier for event detection. To reduce the geolocation error, this paper proposes a noise toponym filtering algorithm (HIST) based on the **hi**erarchical **s**tructure of **t**oponyms. HIST analyzes the hierarchical structure of toponyms mentioned in the event cluster, taking the highly frequent city-level locations as the coarse-grained locations for events. By comparing the hierarchical structure of the toponyms within the cluster against those of the coarse-grained locations of events, HIST filters out noisy toponyms. To further improve the geolocation accuracy, we propose a **fi**ne-grained pseudo **t**oponyms generation algorithm (FIT) based on the output of HIST, and combine generated pseudo toponyms with filtered toponyms to locate events based on the geographic center points of the combined toponyms. Extensive experiments are conducted on the Chinese dataset constructed in this paper and another public English dataset. The experimental results show that the proposed method is superior to the state-of-the-art baselines.

[*] Corresponding author. Email address: kitesmile@126.com

# 1 Introduction

A social event is a specific occurrence at a specific time and place that attracts attention and generates news traffic [1]. For instance, for the arrest of the founder of the social media platform "Telegram" in France that happened in August 2024, the experiencer was the first source to report and update information about the explosion. Detecting the location of such an event is crucial for understanding event-related information, as it provides valuable insights for timely response. The geolocation of events can support a variety of applications [2], notably in facilitating prompt emergency response [3] and efficient Disaster Management [4,5], among others.

Current research on event geolocation is primarily centered on two aspects. The first aspect focuses on geolocating events that have already been classified, emphasizing the extraction of location indicators from event-related content to pinpoint the locations where the events occurred [6]. This approach is more efficient when dealing with accurately classified events. Nevertheless, in real-world applications, obtaining such precisely classified events is often challenging. The second aspect tackles this issue by first detecting events within social data and then performing geolocation based on the detection results [7]. This approach exhibits greater adaptability but carries the risk that errors generated during event detection may propagate into the geolocation process, thereby influencing the final geolocation accuracy and posing additional challenges. Given these considerations, this paper primarily focuses on the research of end-to-end event geolocation methods, aiming to address the limitations of the existing approaches and enhance the overall performance in practical scenarios.

However, the existing end-to-end event geolocation methods are hampered by two notable limitations. Firstly, the phenomenon of error propagation poses a significant challenge: errors in the event detection stage are propagated to the subsequent geolocation process, resulting in a compounding effect where the accuracy of geolocation outcomes diminishes as the error in event detection escalates. Secondly, using all location mentions in text for event localization can lead to significant errors. Despite many locations are mentioned in text, not all are relevant to the event's occurrence location [8,9], and significant geolocation errors can result from using all mentioned geographical information for geolocation.

To address these challenges, this paper proposes a novel end-to-end event geolocation method leveraging Hyperbolic space and toponym hierarchies. The proposed method comprises two key components: an event detection module based on Hyperbolic space, and a geolocation module based on the hierarchical structure of toponyms.

Specifically, during the event detection process, a heterogeneous graph is constructed to fully integrate the heterogeneous information on social media. To accurately depict the tree-like structure of social media, the heterogeneous graph is then transformed into a homogeneous graph, which is further projected into the Hyperbolic space to harness its capability of preserving hierarchical relationships. Subsequently, the SoftMax algorithm is adeptly employed to classify events, enabling precise and efficient event detection.

Given the pivotal role of toponyms in geolocation [10,11], we propose a noise toponym filtering algorithm (HIST) for improving the accuracy of event geolocation processes. HIST leverages the hierarchical structure of toponyms to mitigate the geolocation errors that stem from the cumulative propagation of event detection inaccuracies. This algorithm begins by parsing the hierarchical structure of the toponyms extracted from the event cluster, thereby deriving the coarse-grained location of the event. Subsequently, it filters out the noise toponyms that do not belong to the event cluster, based on the coarse-grained location of the event. To further improve the accuracy of geolocation, we propose a fine-grained pseudo toponyms generation algorithm (FIT). FIT aims to generate fine-grained pseudo toponyms to compensate for the lack of fine-grained pseudo toponyms. The generated fine-grained pseudo-toponyms are integrated with the filtered toponyms to achieve precise event geolocation by calculating the geographical centroid of these toponyms.

The main contributions of this paper are as follows:

- We propose a noise toponym filtering algorithm (HIST) based on the hierarchical structure of toponyms, which considers the hierarchical structure of toponyms and the matching degree of event cluster coarse-grained locations to filter out noisy location information.
- We propose a fine-grained toponym generation algorithm (FIT) that creates fine-grained pseudo-toponyms based on the extracted toponyms from event clusters. Following rigorous validation and selection, these pseudo-toponyms are integrated with the extracted toponyms for the geolocation process.
- We propose a novel end-to-end event detection and geolocation method leveraging Hyperbolic space and toponym hierarchies. Extensive experiments are conducted on both Chinese and English datasets to verify the performance of the proposed method, and the experimental results demonstrate that the it outperforms state-of-the-art baselines.

## 2 Related Work

In this section, we explore the foundational literature that supports our research, concentrating on two pivotal domains: event detection and event geolocation.

**Event detection**

Event detection aims to identify and extract significant events from large datasets for various applications. Common event detection methods can be broadly classified into three distinct categories: rule-based methods [12,13], statistical probability-based methods [14,15], and machine learning-based method [16,17].

Rule-based methods use predefined rules to spot event characteristics, relying on domain expertise. For instance, Kilicoglu et al. [18] created an event dictionary for detecting biological events, while Alhalabi et al. [19] set rules to flag potential terrorism. Despite their simplicity, these methods need regular updates to stay effective, reflecting the evolving nature of events and data.

Statistical and probability-based methods analyze event characteristics by identifying significant patterns in data. For instance, Chierichetti et al. [20] used keyword statistics on social media to find event themes, while Chen et al. [21] applied Markov chains and Bayesian networks for real-time event extraction in water systems. These methods are interpretable and intuitive but rely heavily on the data's quantity and quality for accuracy.

Machine learning-based method identify events by learning patterns from data. For instance, Zhang et al. [22] used Deep Belief Networks (DBN) and Long Short-Term Memory networks (LSTM) to detect traffic-related events. To make more comprehensive use of information from different elements of social media, scholars have proposed methods that utilize heterogeneous information network graphs, such as the Pairwise Popularity Graph Convolutional Network (PP-GCN) by Peng et al [23]. PP-GCN effectively captures and propagates event-related knowledge across heterogeneous information networks by exploiting the semantic relationships established through meta-paths.

**Geolocation**

The objective of event geolocation is to accurately pinpoint the geographical location where an event occurs, to provide precise spatial information support for decision-making systems. Common geolocation methods can be categorized into three types: location oriented-based methods [24], aggregation-based methods [25,26], and probability-based methods [27,28].

Location oriented-based methods transforms the problem of event geolocation into a matter of determining whether an event occurs within a specific region. Such methods first collect tweets from the target area based on given geographical information, and then realize event geolocation by detecting events that occur within that region. For instance, Stefanidis et al. [29] monitored Cairo's Tahrir Square vicinity by collecting tweets from a 10-km radius with location-related hashtags. This facilitated identifying events linked to the geographic hotspot.

Aggregation-based methods combine the locations of geographical entities to find trends or central locations of events. In the process of geolocation, they find event coordinates by calculating the centroid of associated messages' latitude and longitude. In Giridhar et al.'s method [30], location information from tweet texts and user profiles is parsed into latitude and longitude coordinates, and the central point of these coordinates is then employed to locate the event. The aggregation-based methods simple and fast, but can be less accurate with uneven data or noise.

Probability-based methods infer the location of events by establishing probability models. For instance, Ozdikis et al. [28] integrated the credibility scores of different spatial features using the Dempster Shafer theory to estimate the location of events. Furthermore, Ozdikis et al. [31] have also developed a method based on kernel density estimation, which assumes that tweets mentioning the same words are likely to be geographically close. By analyzing the spatial distribution of words in tweets, it predicts the probable geographic locations of new tweets that mention the same words. The probability-based methods are generally influenced by data quality and often have lower computational efficiency compared to other types of methods.

## 3 Problem formulation

To facilitate readers' understanding of the proposed method, this section provides an overview of the problem formulation and definitions of the key terms used in this paper.

A social event is a collection of messages discussing the same real-world event, messages are defined as social content posted by the user. We assume that different social events are independent of each other, i.e., there is no message that related to more than one set of events.

Event geolocation aims to locate the geographical location of events from social media data. This paper breaks down the event geolocation problem into two processes: event detection and geolocation. The aim of event detection is to acquire classified events, whereas the objective of geolocation is to pinpoint the location of each individual event.

Specifically, event detection can be regarded as a classification problem, where each event is considered as an independent class. The input for event detection is a set of messages related to a specific event, which are aggregated into a dataset. The output is the classification results, that is, a series of event clusters, each cluster containing multiple messages related to the same event. The process can be formally expressed as follows:

$$f_d(M;\theta_d) = E \tag{1}$$

where $f_d(\cdot)$ represent the event detection function, $M$ represents the input dataset, $\theta_d$ represents the parameters used in the event detection process, $E = \{e_0, \cdots e_{i-1}, \cdots e_{|N|}\}$ represents the set of output event classes, $e_i$ represents a certain event class, and $|N|$ represents the number of event classes

This paper breaks down the geolocation process into three main steps: toponyms extraction, geocoding, and centroid calculation. The input for this process is the event clusters output from the event detection stage, and the output is a collection of latitude and longitude points corresponding to each event class. The specific process can be formally expressed as follows:

$$f_g(E;\theta_g) = L \tag{2}$$

where $f_g(\cdot)$ represent the geolocation function, $E$ is the set of event classes output by the event detection process, $\theta_g$ represents the parameters used in the geolocation process, and $L = \{l_0, \cdots, l_{i-1}, \cdots l_{|E|}\}$ is the set of event locations. $l_i$ within it represents a latitude and longitude point.

To provide a comprehensive overview of the proposed method, the definitions of some key terms used specifically in this paper are provided below. The notations used in this paper are presented in Table 1.

**Definition 1 (Heterogeneous information networks):** The heterogeneous information network integrates diverse social elements for a comprehensive understanding of social data interactions and dependencies. This paper uses a heterogeneous information graph $\mathcal{G} = (\mathcal{V}, \mathcal{E})$ to represent it, where $\mathcal{V}$ represents the set of nodes, and $\mathcal{E}$ represents the set of edges. This graph consists of message nodes, word nodes, and user nodes, which in turn represent posts, words mentioned in text content, and individuals involved in social interactions. It also includes publish, contain, and interact edges. The publish edge indicates the action of a user publishing a message on social media platform, the contain edge indicates the words are mentioned in a message, and the interact edge represents the interaction between users and messages. In this paper, we use the mention relationship, which indicates that a user mentioned another user in a message, to create the interact edge.

**Definition 2 (Homogeneous message graph):** The homogeneous message graph is transformed from the heterogeneous information graph. In the field of social media, the mechanism of user mentions and shares is crucial for information dissemination, which makes social networks present a tree like structure. Considering this, we designate message nodes as the sole node type within the homogeneous message graph. Homogeneous message graph can be represented as $G=(X,A)$, $X$ represents the node features, $A$ represents the adjacency matrix between nodes. The mapping procedure from heterogeneous information graphs to homogeneous message graphs is detailed in Section 3.3.

**Definition 3 (Geographical hierarchical structure):** Places usually have a certain geographical hierarchy, and each geographical entity has its specific geographical hierarchy. The geographic hierarchy provides an effective way to conceptualize the subordinate relationships between locations, where each level represents a specific geographic area, typically gradually subdivided from larger administrative regions into smaller geographic units [32]. Taking China as an example, starting at the highest level, the country itself, they are gradually subdivided downwards into provinces, cities, districts, counties, townships, streets, roads, and so on, to specific locations or landmarks. Usually, for a certain toponym, we can get its geographical hierarchical structure with the help of toponymic dictionary or map tool. For example, for the toponym "West Lake", we can get "China - Zhejiang Province - Hangzhou City - West Lake District - West Lake Street - Manjuelong Road - West Lake Scenic Area".

Table 1 Glossary of notations.

| Notation | Description |
| --- | --- |
| $M$ | The input dataset |
| $f_d(\cdot); f_g(\cdot)$ | Event detection function; geolocation function |
| $\theta_d; \theta_g$ | Event detection parameters; geolocation parameters |
| $e; E$ | A social event; Set of events |
| $l; L$ | Location of an event; location set of events |
| $\mathcal{G}$ | Heterogeneous information graph |
| $\mathcal{V}; \mathcal{E}$ | The set of nodes, edges in $\mathcal{G}$ |
| $G$ | Homogeneous message graph |
| $X; A$ | The node features, adjacency matrix of $G$ |
| $m; w; u$ | Message node; Word node; User node |
| $H$ | Hyperbolic embeddings |
| $\mathcal{L}$ | Cross-entropy loss |

## 4 Proposed Methods

  This section presents detail introduction of GTOP. Section 4.1 provides an overview of it, while Section 4.2 and Section 4.3 detail the components of it.

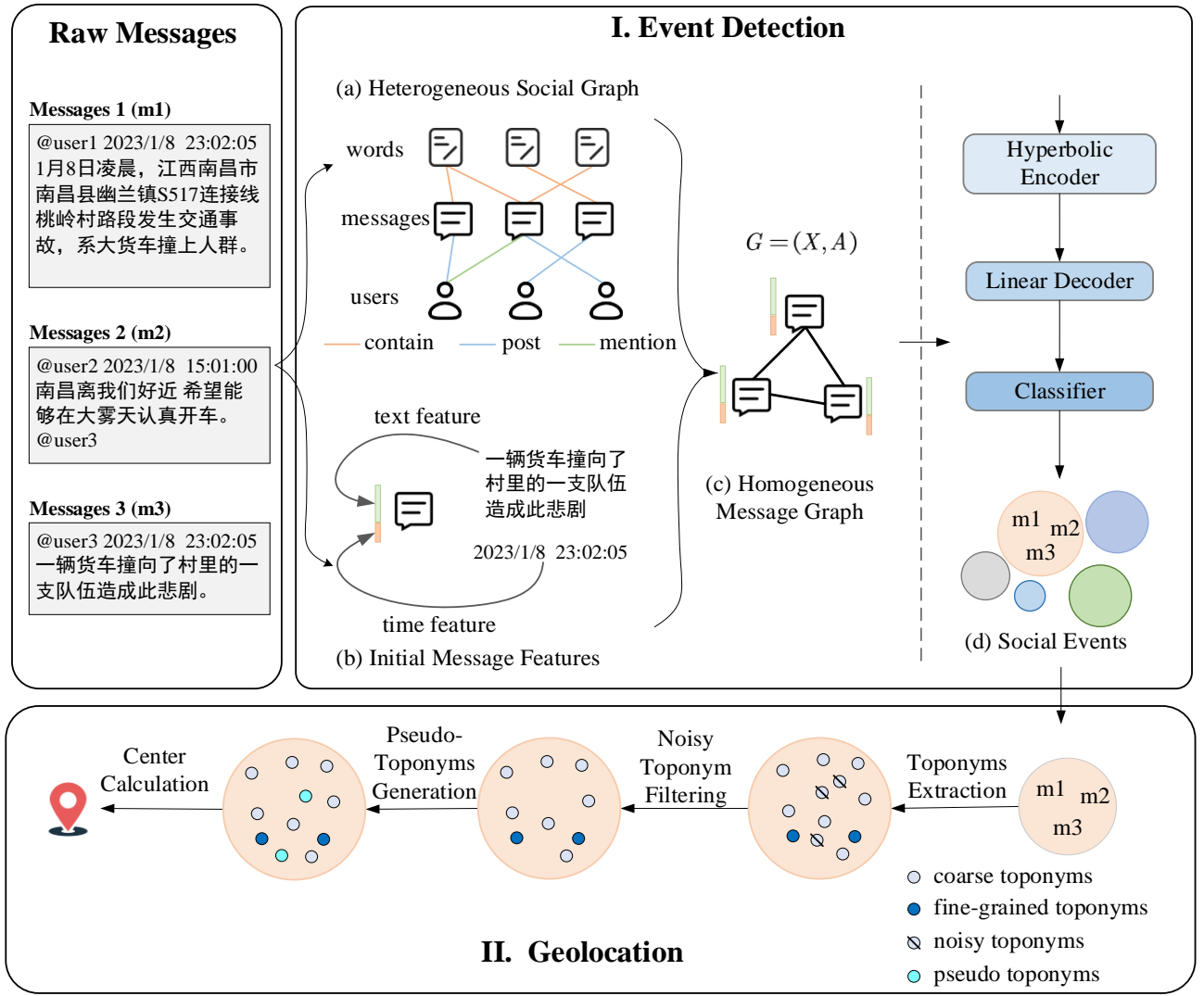

Figure 1 The architecture of GTOP. Different colors in (d) denote distinct event clusters, each of which contains a number of messages.

## 4.1 Overview

In this paper, we propose a novel end-to-end event geolocation method leveraging Hyperbolic space and toponym hierarchies, which follows the process of "event detection-geolocation". The structure of the proposed method is shown in Figure 1.

As Shown in Figure 1, the proposed event geolocation method mainly consists of two parts: event detection and geolocation. In the event detection module, we first model social media data as a heterogeneous information graph and map it onto a homogeneous message graph. Unlike traditional neural network models most of who normally learn node features in Euclidean Space, we map node features into a Hyperbolic space for learning, thereby detecting the occurrence of events more accurate. In the geolocation module, we estimate events' locations by identifying the toponyms mentioned in the messages and calculating the centroid of the coordinates of them. Specifically, to improve the geolocation accuracy, we proposed two algorithms to filter out the noisy toponyms and complement fine-grained location features of the event.

Ultimately, utilizing Baidu Maps, we geocode event-related toponyms to extract latitude and longitude. We then obtain the event's location by calculating the centroid of these coordinates.

## 4.2 Event Detection

The event detection module is structured around three integral phases: the heterogeneous social message modeling, the homogeneous message graph construction and the event detection in Hyperbolic space. The illustration of the first two phases is presented in Figure 2.

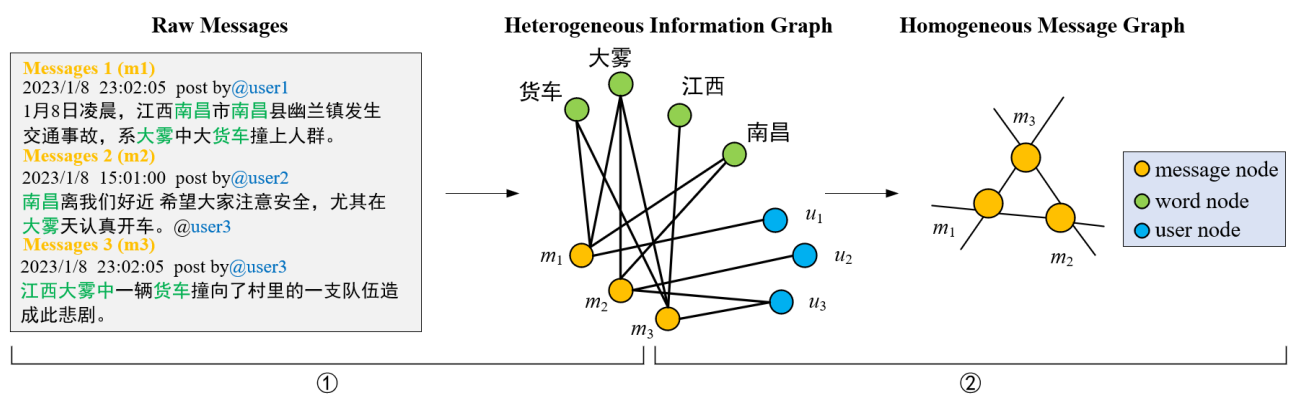

Figure 2 An example of constructing homogeneous message graphs from raw messages. ① illustrates the process of constructing a heterogeneous information graph from raw messages. ② illustrates the process of constructing a homogeneous message graph.

### 4.2.1 Heterogeneous Social Message Modeling

The social messages that are most closely related to event geolocation are temporal, location, and semantic information. Inspired by Peng et al. [33], to make full use of the available information, we model social messages as a heterogeneous information graph that contain user nodes, word nodes and message nodes, which are denoted as *w*, *u*, and *m* respectively. The process of constructing a heterogeneous information graph from raw messages is presented in Figure 2.

For instance, through extracting words that have a high frequency of occurrence and are related to geographical entities, we can obtain various types of nodes for message $m_1$, including message node $m_1$, user node $u_1$, and word nodes "南昌", "货车", and "大雾". This process is repeated for all messages to build the whole heterogeneous information graph. For repeated nodes, we only retain one node and keep all its connected edges. For example, if user $u_3$ is mentioned in messages $m_1$ and $m_2$, we only create one $u_3$ node in the graph, but we retain the edges connecting $u_3$ to both $m_1$ and $m_2$.

In order to integrate both semantic and temporal information, we undertake the construction of initial message vectors. Specifically, semantic information is extracted utilizing the spaCy library, which provides robust tools for linguistic analysis. Concurrently, timestamps associated with each message are transformed into OLE dates (Object Linking and Embedding Date format). This transformation entails decomposing the timestamp into its fractional and integer components, which are then arranged into a two-dimensional vector, serving as temporal features. By concatenating these semantic features with the temporal features, we create a comprehensive initial representation for each message node within the heterogeneous information graph.

### 4.2.2 Homogeneous Message Graph Construction

Since GTOP focus on learning the correlations between messages, we transform a heterogeneous information graph into a homogeneous message graph, where only messages serve as nodes. The mapping process is as Equation (3),

$$A_{i,j} = \min\left\{\left[\sum_k W_{mk} \cdot W_{mk}^T\right]_{i,j}, 1\right\}, \quad k \in \{o, e, u\} \quad (3)$$

where $A_{i,j}$ represents the element in the *i*-th row and *j*-th column of matrix *A*, *A* is the adjacency matrix of the homogeneous message graph. *m* denotes the message node, and *k* represents the other two node types. $W_{mk}$ is a submatrix of the adjacency matrix of the heterogeneous information graph, containing rows of type *m* and columns of type *k*.

From Equation (3), we can see that if messages $m_i$ and $m_j$ link to same nodes, $[W_{mk} \cdot W_{mk}^\top]_{i,j}$ will be greater than or equal to 1, hence $A_{i,j}$ will be equal to one. That means if two message nodes have at least one common neighbor node, then we create an edge between them in the homogeneous message graph.

The process of constructing a homogeneous message graph from the heterogeneous information graph is given in Figure 2. We can see that, $m_1$ and $m_2$ can be connected through the word node "南昌", $m_2$ and $m_3$ can be connected through the user node user $u_3$, and $m_1$ and $m_3$ can be connected through the word node "大雾".

### 4.2.3 Event Detection in Hyperbolic Space

Most event detection methods approach the problem as a classification task within Euclidean space [23,34,35]. However, social media data typically exhibits a tree-like or hierarchical structure[36], and Euclidean space cannot accommodate its exponential data growth. Compared to Euclidean space, Hyperbolic space can more effectively cope with the exponential growth of social media data and can embed tree-like structures without loss. Therefore, we choose to conduct event detection in hyperbolic space.

Specifically, our node embeddings are defined in Euclidean space. During the training process, when updating node features, we map them into hyperbolic space for the update. However, vectors in Euclidean space cannot be directly translated into hyperbolic space; they must undergo a specific mathematical transformation to achieve this mapping. To address this challenge, we have established a mapping between hyperbolic space and Euclidean space utilizing exponential and logarithmic functions.

For instance, for embeddings $α$ in Euclidean space, given the constant negative curvature *c* of a hyperbolic space and the origin point *o* of a hyperbolic space, using Equation (4), we can obtain their representation $α′$ in hyperbolic space. Subsequently, the node features are updated within the hyperbolic space. After the feature updates within the hyperbolic space, we proceed to use a linear decoder to map the updated features back into Euclidean space. For embeddings $β$ in hyperbolic space, using Equation (5), we can map them back to Euclidean space to obtain $β′$, thereby ensuring compatibility with conventional machine learning frameworks.

$$\alpha' = exp_o^c(\alpha) = tanh(\sqrt{c}\|\alpha\|)\frac{\alpha}{\sqrt{c}\|\alpha\|} \quad (4)$$

$$\beta' = log_o^c(\beta) = artanh(\sqrt{c}\|\beta\|)\frac{\beta}{\sqrt{c}\|\beta\|}$$

(5)

In the final step, the node feature vectors are input into a softmax classification layer, which facilitates the prediction and assignment of appropriate labels.

### 4.3 Geolocation

Regarding the event geolocation process, we represent the event location as a latitude and longitude point, not a toponym, so we need to geo-code it with the help of a toponym dictionary or a mapping application, and in this paper, we use Baidu map.

### 4.4.1 Toponyms extraction

For clusters of output events, we need to extract the geographic information in them, mostly in the form of toponyms, so we use the named entity tool to extract the toponyms in the messages.

In the process of Chinese data processing, we chose jieba as a tool for toponymic extraction. Jieba is a widely used Chinese word-splitting tool, which has high accuracy in Chinese text processing, and is suitable for a variety of tasks such as lexical annotation, keyword extraction, and named entity recognition. jieba employs a keyword extraction technique based on the TF-IDF algorithm, as well as a Hidden Markov Model (HMM) based lexical annotation method. In addition, jieba supports the flexible configuration of custom dictionaries, which makes it adaptable to a variety of different Chinese text processing needs. For the processing of English data, we used spaCy as a toponymic extraction tool. spaCy is an advanced Natural Language Processing (NLP) library designed for large-scale information extraction.

### 4.4.2 Noisy toponym filtering

Since there is always some error in the event detection work, messages from other clusters may be included within a certain cluster, and the extracted toponyms will have a large error at this time. To solve this problem, we can use the hierarchical relationship of toponyms for outlier filtering to reduce the influence of noisy data.

With the help of Baidu map, we can obtain information about the toponym hierarchy in Chinese, and the granularity can be divided into "province-city-district/county-township/street-village/community-street" from coarse to fine (some hierarchies do not necessarily exist). For a collection of toponyms (extracted from a certain event cluster), we can determine the representative toponyms at each level by counting the occurrences of each toponym and selecting the one with the highest frequency. For instance, if in a set of toponyms, "Henan" appears 500 times, "Jiangxi" appears 400 times, and "Guangdong" appears 300 times, then "Henan" would be the representative toponym at the provincial level. We have defined the concept of a "cluster toponym chain," which is a sequence connected by representative toponyms of different levels according to the precision of geographical location. For a specific set of toponyms, its cluster toponym chain is fixed and unchanging.

After obtaining the cluster toponym chain, we need to parse the hierarchical information of each toponym in the collection and compare it with the cluster toponym chain for verification. If a toponym does not match the cluster toponym chain, it is considered a noisy toponym. When making matches, in most cases, we only use the "province-city" two-level toponym chain for matching, which means that if the province and city information parsed from the toponym matches the cluster toponym chain, then the toponym is valid. For events that require higher precision location, we will use a more detailed toponym chain, such as including "district/county" or even "town/street" and other finer levels.

For instance, as illustrated in Figure 2, for the toponym "West Lake", we can get information about its hierarchical toponym. If it does not match the cluster toponym chain, the toponym "West Lake" is considered as noisy toponym and is deleted from the toponym list. This method can ensure that we accurately identify and eliminate noisy toponyms, thereby improving the quality and practicality of toponym data.

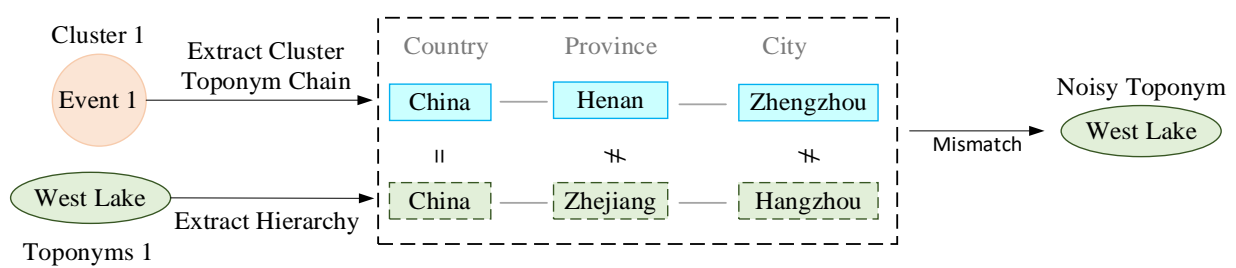

Figure 2 An example of outlier filtering

### 4.4.3 Pseudo-toponyms generation

In order to improve the performance of geolocation, introducing some pseudo toponyms in addition to the extracted toponyms is a feasible approach. Specifically, by counting the frequency of toponym occurrences in the event clusters, we can identify the toponyms that have a high number of occurrences under each stratum. Subsequently, these toponyms are linked according to their hierarchical structure. Usually, for an event cluster, we can at least extract toponym information at the "province-city" level. If there are more detailed hierarchical information, we connect them in turn. In the end, we construct multiple toponyms that conform to the toponym hierarchy, which are spliced from the extracted toponyms in the cluster, and are called pseudo-toponyms because they do not appear directly in the event text.

---

**Algorithm 1: Hyperbolic Event Geolocation Mapper**

**input** : Set of messages: $M$
**output**: Set of event locations: $L$

1. **for** $m_i \in M$ **do**
2. $\quad \mathcal{G} = \text{AddGraphNodes}(m_i)$
3. $G = \text{HomogeneousMessageGraph}(\mathcal{G})$
4. $H = \text{MLPEncoder}(G)$
5. $Z = \text{LinearDecoder}(H)$
6. $E = \text{Classify}(Z)$
7. **for** $e \in E$ **do**
8. $\quad \text{locations} = \text{ExtractToponym}(e)$
9. $\quad \text{max\_location} = \text{MaxLocations(locations)}$
10. $\quad \text{pseudo\_location} = \text{max\_location.district} + \text{max\_location.township} + \text{max\_location.village} + \text{max\_location.street}$
11. $\quad \text{locations} = \text{add(pseudo\_location)}$
12. $\quad$ **for** *location in locations* **do**
13. $\quad\quad$ **if** $\text{location.province} \neq \text{max\_location.province}$ or $\text{location.city} \neq \text{max\_location.city}$ **then**
14. $\quad\quad\quad \text{locations} = \text{remove(location)}$
15. $\quad \text{coords} = \text{Geocoding(locations)}$
16. $\quad l_i = \text{Computing Center(coords)}$

---

In the process of generating pseudo-toponyms, some accuracy problems may be encountered as these toponyms are formed by directly splicing toponym information from different hierarchies. Especially when the toponyms of different tiers point to geographically different locations, the direct use of these pseudo-toponyms without validation may introduce geolocation errors. To address this issue, similar to the noisy toponym filtering approach, we first parsed the generated pseudo-toponyms using a mapping application to determine their coarse-grained geographic hierarchical structure. Subsequently, we analyze the parsed hierarchical structure in comparison with the toponym chains of the event clusters. Only if the two match, we include the pseudo-toponym in the geolocation process.

### 4.4.4 Center calculation

After the previous process, for a certain event cluster, we obtain a number of toponyms, including those extracted directly from the text as well as generated pseudo-toponyms. In order to match these toponyms with the actual geographic locations, the toponyms need to be converted into specific geographic coordinates. We used Baidu map for the geocoding work.

After obtaining the latitude and longitude information, we need to determine a representative center location from these scattered coordinate points. We used KMeans for clustering, the core idea of which is to divide the data points into clusters so that the points within the clusters are as similar as possible and the points between the clusters are as different as possible. The algorithm starts by randomly selecting K data points as the initial cluster centers, and then assigns each point to the nearest cluster based on the distance from each data point to each cluster center. Subsequently, the algorithm updates the center point of each cluster, usually taking the mean of all points in that cluster as the new center. This process is repeated until the change in the cluster centers is very small or a preset number of iterations is reached, at which point the algorithm stops and the final cluster centroids obtained are the clustering results. In our application, we set the cluster to 1, i.e., we look for the centroid of all coordinate points, which can be considered as the geographic location of the event. The pseudo-code for our model is shown in Algorithm 1.

## 5 Experiments

### 5.1 Datasets

Currently, Chinese event geolocation datasets are relatively limited. To fill this gap, we constructed our own Chinese geolocation dataset based on the microblogging platform, which contains 8023 manually tagged microblogs related to 37 event categories after filtering out duplicates and meaningless microblogs.

The process of building the dataset is shown in Figure 4. First, we collect event-related keywords and crawl the raw data on the microblogging platform. For the crawled raw microblog data, we first need to go through cleaning: (1) filter duplicate microblogs: for duplicate microblogs, we keep only the one with the earliest publication time. (2) filter irrelevant microblogs: delete microblogs that contain event keywords but are not really related to the event. (3) filter out low-information microblogs: we remove microblogs with low information, such as microblogs that are too short. (4) filter outdated microblogs: delete microblogs that were published long after the event occurred. Finally, we label each event class ground truth.

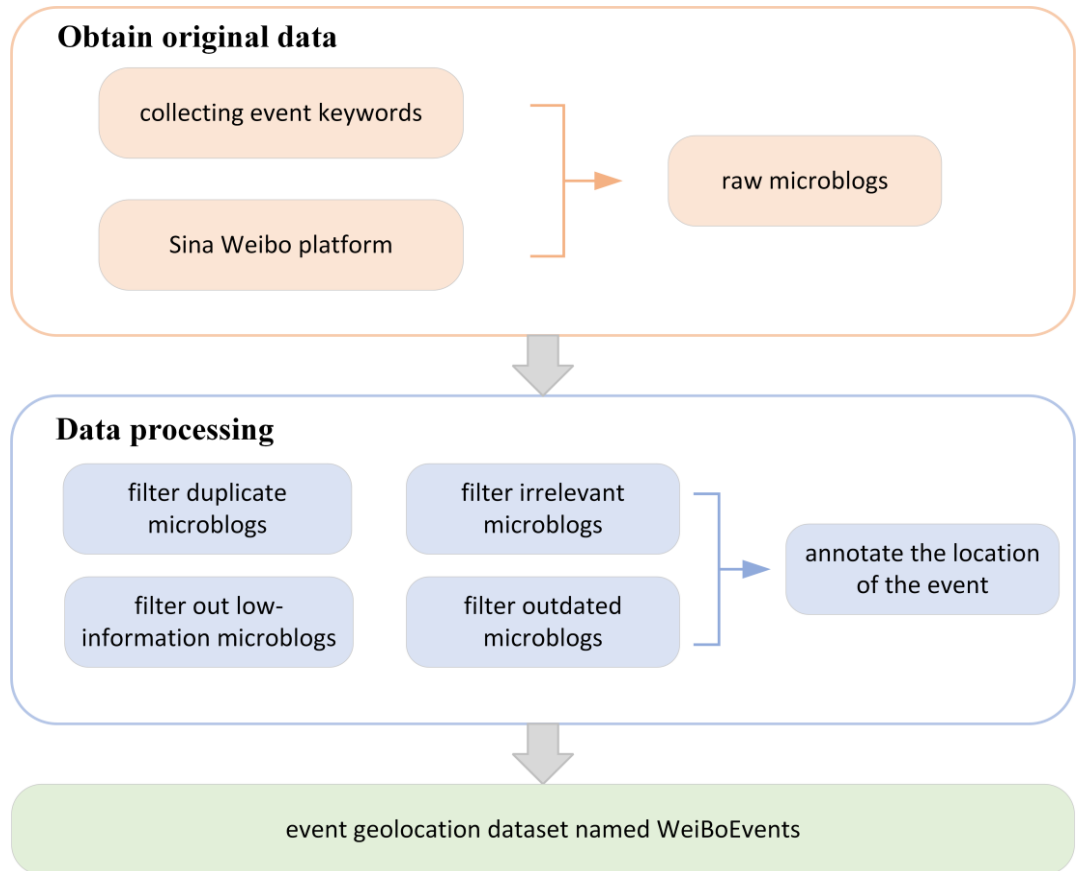

Figure 4    The dataset construction route

Ultimately,Our experiments were conducted on two datasets.:

IDRISI: This dataset comprises approximately 205,000 manually annotated tweets for location mention recognition. It includes tweets related to 19 different types of disaster events. We extracted a subset of 8128 tweets belonging to 8 event categories for event geolocation.

WeiBoEvents: Our Chinese event geolocation dataset constructed based on the Weibo platform contains 8,023 unique Weibo posts related to 37 event categories after filtering out duplicates and irrelevant content.

Ultimately,Our experiments were conducted on two datasets.:

IDRISI: This dataset comprises approximately 205,000 manually annotated tweets for location mention recognition. It includes tweets related to 19 different types of disaster events. We extracted a subset of 8128 tweets belonging to 8 event categories for event geolocation.

WeiBoEvents: Our Chinese event geolocation dataset constructed based on the Weibo platform contains 8,023 unique Weibo posts related to 37 event categories after filtering out duplicates and irrelevant content.

The comparison of the two datasets is shown in Table 2, and their ground truth is illustrated in Figures 5-6.

Table 2　Comparative analysis of datasets

| dataset | Number of Messages | Number of Events | Average Number of Messages |
| --- | --- | --- | --- |
| IDRISI | 8128 | 9 | 903 |
| WeiBoEvents | 8023 | 37 | 216 |

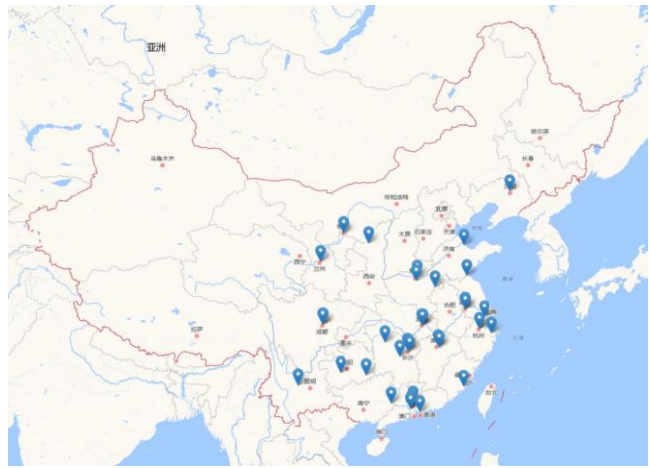

Figure 5 Geographical distribution of ground truth in WeiBoEvents dataset

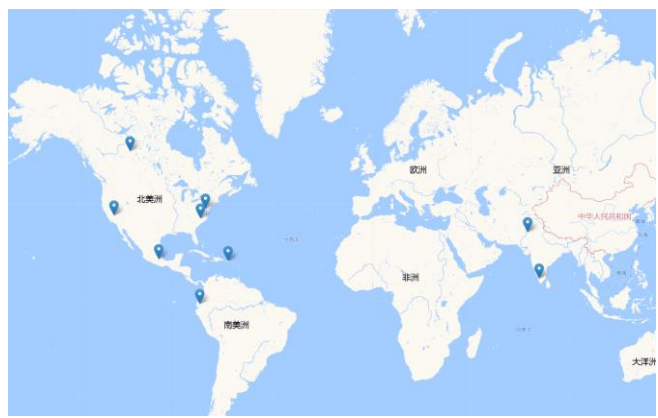

Figure 6 Geographical distribution of ground truth in IDRISI dataset

## 5.2 Baseline

We compared the GTOP with the following methods. The baselines are:

(1) SiameseUniNLU: A universal natural language understanding model designed to handle various tasks by using prompts tailored to specific tasks and employing a Pointer Network for span extraction.

(2) RexUniNLU[37]：RexUniNLU is a Universal Information Extraction (UIE) model based on recursive methods and explicit schema guidance. We take advantage of its powerful entity extraction capabilities for toponymic identification.

(3) RaNER[38]: A named entity recognition model with a Transformer-CRF architecture using StructBERT as a pre-trained model, combined with the use of relevant sentences recalled by an external tool as additional context.

(4) GTOP--: A variant of GTOP,which lacks "Outlier filtering" and "Generation of fine-grained pseudo-toponyms" compared to GTOP.

Due to the limited availability of complete event geolocation methods, we uniformly adopted the classification results of the proposed GTOP model.

## 5.3 Evaluation Metrics.

To evaluate the performance of all models, we use error distances to evaluate the quality of geolocation, More in detail, we use the mean error distance and median error distance.

The error distance between the estimated geographic coordinates and the true geographic coordinates is calculated using Haversine's formula:

$$dis\tan ce = 2r\sin^{-1}\left(\sqrt{\sin^2\left(\frac{\hat{g}_{long} - g_{long}}{2}\right) + \cos(\hat{g}_{long})\cos(g_{long})\sin^2\left(\frac{\hat{g}_{lat} - g_{lat}}{2}\right)}\right)$$

(6)

$R$ is the radius of the Earth. $(\hat{g}_{long}, \hat{g}_{lat})$ and $(g_{long}, g_{lat})$ are the estimated event geographic coordinates and the real event geographic coordinates, respectively. In order to further enrich the study of the performance of our approach.

We use $ACC_d$ to denote the accuracy of the method, Formally,

$$ACC_d = \frac{|\{e|distance(e) \leq d\}|}{N} \qquad (7)$$

$N$ is the number of events.

### 5.4 Result

Our ultimate goal is to detect the occurrence of an event from social media data and locate the location of its occurrence. Table 3-4 shows the experimental results of the GTOP model and the baseline model on our dataset. Overall, the GTOP model outperforms the baseline model on every metric.

Table 3　The error distance in WeiBoEvents dataset

| model | Mean | Median |
|---|---|---|
| RaNER | 296.68 | 271.19 |
| SiameseUniNLU | 265.14 | 274.42 |
| RexUniNLU | 293.40 | 274.99 |
| GTOP-- | 292.92 | 318.10 |
| GTOP | 223.46 | 209.98 |
| GTOP++ | **208.88** | **133.55** |

Table 4　The $ACC_d$ in WeiBoEvents dataset

| model | $ACC_{d=100}$ | $ACC_{d=200}$ | $ACC_{d=300}$ | $ACC_{d=400}$ |
|---|---|---|---|---|
| RaNER | 18.92% | 40.54% | 51.35% | 72.97% |
| SiameseUniNLU | 13.51% | 37.84% | 56.76% | 81.08% |
| RexUniNLU | 13.51% | 35.14% | 51.35% | 75.68% |
| GTOP-- | 16.22% | 35.14% | 48.65% | 75.68% |
| GTOP | 24.32% | 48.65% | **78.38%** | **86.49%** |
| GTOP++ | **32.43%** | **59.46%** | 67.57% | **86.49%** |

It is worth noting that the GTOP-- variant performs weakly without the "Outlier filtering" and "Generation of fine-grained pseudo-toponyms" steps, mainly constrained by the noise data introduced during event detection. However, once the noise filtering mechanism is incorporated (i.e., the GTOP model), its performance significantly improves, achieving optimal results.

Table 5   The error distance in IDRISI dataset

| model | Mean | Median |
|---|---|---|
| RaNER | 3028.64 | 3262.23 |
| SiameseUniNLU | 3262.09 | 2832.43 |
| RexUniNLU | 3606.35 | 2374.68 |
| GTOP-- | 3531.98 | 3883.51 |
| GTOP | 406.54 | **238.74** |
| GTOP++ | **356.84** | 255.12 |

This phenomenon is even more pronounced in the IDRISI dataset. Because IDRISI involves worldwide geolocation of events, the error transmission effect is even more severe. It can be clearly seen that the effect of GTOP is significantly improved compared to other baselines. This indicates that the GTOP model has excellent ability to resist the error transfer effect. For the complete event geolocation task, it has more powerful performance.

In our geolocation phase, the main reliance lies in the recognition of toponyms. Currently, our primary tools for toponym recognition are the jieba library (for Chinese) and the spaCy library (for English); however, compared to the latest named entity recognition methods, their effectiveness is relatively limited. In order to comprehensively assess this issue, we have introduced the GTOP++ model and utilized the optimal baseline model to replace the jieba library and the spaCy library for toponym recognition. The results indicate a further improvement in performance with the GTOP model.

# Discussion

However, previous research has overlooked one key aspect: social media data typically exists in the form of user information, posts, comments, shares, and so on, with complex parent-child relationships between them, forming a tree-like or hierarchical structure [17]. The exponential growth of data, driven by an increasing number of users and frequent interactions, presents a formidable challenge to conventional data processing models. These models typically project data into Euclidean space, which grows polynomially—a rate that fails to match the exponential expansion of social media data. As a result, existing models that map social media data into Euclidean space are unable to effectively express or fully capture its tree-like structure [39]. Compared to Euclidean space, hyperbolic space offers notable advantages in detecting and locating events using social media data. It naturally expresses hierarchical structures inherent in social media, accommodating exponential data growth more effectively. The distance metric in hyperbolic space aligns with power-law distributions common in social media, enabling accurate modeling of scale-free networks. Additionally, it can embed tree structures losslessly, crucial for tracing information propagation paths. These features make hyperbolic space a powerful tool for social media analysis.

# 6 Conclusion

In this study, we propose a hyperbolic event geolocation model, GTOP, which aims to discover events and extract the location of events through social media data. The model focuses on the tree structure of social media data, using hyperbolic space to enhance semantic and structural information. The hierarchical relationship of toponyms is used to enhance the localisation effect. We empirically demonstrate the superiority of GTOP compared to baselines through experiments. In future work, we will also consider applying our model to dynamic social event geolocation.

**ACKNOWLEDGMENTS**

This research was funded by National Natural Science Foundation of China (62272163)

# Reference

[1]   Zhang Y, Szabo C, Sheng Q Z, et al. SNAF: Observation filtering and location inference for event monitoring on twitter[J]. World Wide Web, 2018, 21(2): 311-343.

[2]   Yu M, Bambacus M, Cervone G, et al. Spatiotemporal event detection: a review[J]. International Journal of Digital Earth, 2020, 13(12): 1339-1365.

[3]   Simon T, Goldberg A, Adini B. Socializing in emergencies - A review of the use of social media in emergency situations[J]. 2015, 35(5): 609-619.

[4]   Huang X, Li Z, Wang C, et al. Identifying disaster related social media for rapid response: a visual-textual fused CNN architecture[J]. International Journal of Digital Earth, 2020, 13(9): 1017-1039.

[5]   Karimiziarani M, Moradkhani H. Social response and disaster management: Insights from twitter data

assimilation on hurricane ian[J]. International journal of disaster risk reduction, 2023, 95: 103865.

[6] Ying Y, Peng C, Dong C, et al. Inferring event geolocation based on Twitter[C]//Proceedings of the 10th international conference on internet multimedia computing and service, nanjing, china, august 17-19, 2018. ACM, 2018: 26:1-26:5.

[7] George Y M, Karunasekera S, Harwood A, et al. Real-time spatio-temporal event detection on geotagged social media[J]. J. Big Data, 2021, 8(1): 91.

[8] Middleton S E, Kordopatis-Zilos G, Papadopoulos S, et al. Location extraction from social media: Geoparsing, location disambiguation, and geotagging[J]. ACM Trans. Inf. Syst., 2018, 36(4): 40:1-40:27.

[9] McKitrick M K, Schuurman N, Crooks V A. Collecting, analyzing, and visualizing location-based social media data: review of methods in GIS-social media analysis[J]. GeoJournal, 2023, 88(1): 1035-1057.

[10] Qi Y, Zhai R, Wu F, et al. CSMNER: a toponym entity recognition model for chinese social media[J]. ISPRS International Journal of Geo-Information, 2024, 13(9): 311.

[11] Zhao Y, Zhang D, Jiang L, et al. EIBC: a deep learning framework for Chinese toponym recognition with multiple layers[J]. Journal of Geographical Systems, 2024, 26(3): 407-425.

[12] Ho S S, Lieberman M, Wang P, et al. Mining future spatiotemporal events and their sentiment from online news articles for location-aware recommendation system[C]//Proceedings of the First ACM SIGSPATIAL International Workshop on Mobile Geographic Information Systems. Redondo Beach California: ACM, 2012: 25-32.

[13] Saeed Z, Abbasi R A, Razzak I, et al. Enhanced Heartbeat Graph for emerging event detection on Twitter using time series networks[J]. Expert Systems With Applications, 2019, 136: 115-132.

[14] Dey N, Sabah Mredula M, Nazmus Sakib Md, et al. A Machine Learning Approach to Predict Events by Analyzing Bengali Facebook Posts[M]//KAISER M S, BANDYOPADHYAY A, MAHMUD M, et al. Proceedings of International Conference on Trends in Computational and Cognitive Engineering: Vol. 1309. Singapore: Springer Singapore, 2021: 133-143.

[15] Zhou S, Kan P, Huang Q, et al. A guided latent Dirichlet allocation approach to investigate real-time latent topics of Twitter data during Hurricane Laura[J]. Journal of Information Science, 2023, 49(2): 465-479.

[16] Hossny A H, Mitchell L. Event detection in twitter: A keyword volume approach[C]//TONG H, LI Z J, ZHU F, et al. 2018 IEEE international conference on data mining workshops, ICDM workshops, singapore, singapore, november 17-20, 2018. IEEE, 2018: 1200-1208.

[17] Huang L, Liu G, Chen T, et al. Similarity-based emergency event detection in social media[J]. Journal of Safety Science and Resilience, 2021, 2(1): 11-19.

[18] Kilicoglu H, Bergler S. Syntactic dependency based heuristics for biological event extraction[C]//Proceedings of the BioNLP 2009 workshop companion volume for shared task, BioNLP@HLT-NAACL 2009 - shared task, boulder, colorado, USA, june 5, 2009. Association for Computational Linguistics, 2009: 119-127.

[19] Alhalabi W, Jussila J, Jambi K M, et al. Social mining for terroristic behavior detection through Arabic tweets characterization[J]. Future Gener. Comput. Syst., 2021, 116: 132-144.

[20] Chierichetti F, Kleinberg J M, Kumar R, et al. Event detection via communication pattern

analysis[C]//ADAR E, RESNICK P, CHOUDHURY M D, et al. Proceedings of the eighth international conference on weblogs and social media, ICWSM 2014, ann arbor, michigan, USA, june 1-4, 2014. The AAAI Press, 2014.

[21] Mao Y, Chen X, Xu Z. Real-time event detection with water sensor networks using a spatio-temporal model[C]//GAO H, KIM J, SAKURAI Y. Database systems for advanced applications - DASFAA 2016 international workshops: BDMS, BDQM, MoI, and SeCoP, dallas, TX, USA, april 16-19, 2016, proceedings: Vol. 9645. Springer, 2016: 194-208.

[22] Zhang Z, He Q, Gao J, et al. A deep learning approach for detecting traffic accidents from social media data[J]. Transportation research part C: emerging technologies, 2018, 86: 580-596.

[23] Peng H, Li J, Gong Q, et al. Fine-grained event categorization with heterogeneous graph convolutional networks[C]//KRAUS S. Proceedings of the twenty-eighth international joint conference on artificial intelligence, IJCAI 2019, macao, china, august 10-16, 2019. ijcai.org, 2019: 3238-3245.

[24] Ozdikis O, Oguztüzün H, Karagoz P. A survey on location estimation techniques for events detected in Twitter[J]. Knowledge and Information Systems, 2017, 52(2): 291-339.

[25] Sankaranarayanan J, Samet H, Teitler B E, et al. TwitterStand: news in tweets[C]//AGRAWAL D, AREF W G, LU C T, et al. 17th ACM SIGSPATIAL international symposium on advances in geographic information systems, ACM-GIS 2009, november 4-6, 2009, seattle, washington, USA, proceedings. ACM, 2009: 42-51.

[26] Sakaki T, Okazaki M, Matsuo Y. Tweet analysis for real-time event detection and earthquake reporting system development[J]. IEEE Transactions on Knowledge and Data Engineering, 2013, 25(4): 919-931.

[27] Giridhar P, Wang S, Abdelzaher T F, et al. Joint localization of events and sources in social networks[C]//2015 international conference on distributed computing in sensor systems, DCOSS 2015, fortaleza, brazil, june 10-12, 2015. IEEE Computer Society, 2015: 179-188.

[28] Ozdikis O, Oguztüzün H, Karagoz P. Evidential estimation of event locations in microblogs using the Dempster-Shafer theory[J]. 2016, 52(6): 1227-1246.

[29] Stefanidis A, Crooks A, Radzikowski J. Harvesting ambient geospatial information from social media feeds[J]. GeoJournal, 2013, 78: 319-338.

[30] Giridhar P, Abdelzaher T F, George J, et al. On quality of event localization from social network feeds[C]//2015 IEEE international conference on pervasive computing and communication workshops, PerCom workshops 2015, st. Louis, MO, USA, march 23-27, 2015. IEEE Computer Society, 2015: 75-80.

[31] Ozdikis O, Ramampiaro H, Nørvåg K. Locality-adapted kernel densities of term co-occurrences for location prediction of tweets[J]. 2019, 56(4): 1280-1299.

[32] Zhao Y, Zhang D, Jiang L, et al. EIBC: a deep learning framework for Chinese toponym recognition with multiple layers[J]. Journal of Geographical Systems, 2024, 26(3): 407-425.

[33] Peng H, Zhang R, Li S, et al. Reinforced, incremental and cross-lingual event detection from social messages[J]. IEEE Transactions on Pattern Analysis and Machine Intelligence, 2023, 45(1): 980-998.

[34] Cao Y, Peng H, Wu J, et al. Knowledge-preserving incremental social event detection via heterogeneous gnns[C]//LESKOVEC J, GROBELNIK M, NAJORK M, et al. WWW '21: The web conference 2021, virtual event / ljubljana, slovenia, april 19-23, 2021. ACM / IW3C2, 2021: 3383-3395.

[35] Ren J, Jiang L, Peng H, 等. From known to unknown: Quality-aware self-improving graph neural network for open set social event detection[C]//HASAN M A, XIONG L. Proceedings of the 31st ACM international conference on information & knowledge management, atlanta, GA, USA, october 17-21, 2022. ACM, 2022: 1696-1705.

[36] Adcock A B, Sullivan B D, Mahoney M W. Tree-like structure in large social and information networks[C]//XIONG H, KARYPIS G, THURAISINGHAM B, et al. 2013 IEEE 13th international conference on data mining, dallas, TX, USA, december 7-10, 2013. IEEE Computer Society, 2013: 1-10.

[37] Liu C, Zhao F, Kang Y, et al. RexUIE: A Recursive Method with Explicit Schema Instructor for Universal Information Extraction[C]//BOUAMOR H, PINO J, BALI K. Findings of the Association for Computational Linguistics: EMNLP 2023, Singapore, December 6-10, 2023. Association for Computational Linguistics, 2023: 15342-15359.

[38] Wang X, Jiang Y, Bach N, et al. Improving Named Entity Recognition by External Context Retrieving and Cooperative Learning[C]//Proceedings of the 59th Annual Meeting of the Association for Computational Linguistics and the 11th International Joint Conference on Natural Language Processing (Volume 1: Long Papers). Online: Association for Computational Linguistics, 2021: 1800-1812.

[39] Yang H, Chen H, Pan S, et al. Dual space graph contrastive learning[C]//LAFOREST F, TRONCY R, SIMPERL E, et al. WWW '22: The ACM web conference 2022, virtual event, lyon, france, april 25 - 29, 2022. ACM, 2022: 1238-1247.